\documentclass{WileyMSP-template}
\usepackage[super]{natbib}
\usepackage[T1]{fontenc}
\usepackage{amsmath}
\makeatletter
\renewcommand\NAT@open{[}
\renewcommand\NAT@close{]}
\renewcommand\NAT@sep{,}
\makeatother

\begin{document}

\pagestyle{fancy}
\rhead{\includegraphics[width=2.5cm]{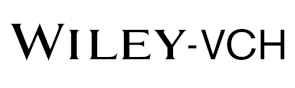}}

\title{Towards Artificial Nerves: Biomimetic Optical-Fiber Tactile Sensing for Robots}

\maketitle

\author{Laura E. Butcher}
\author{Chris J. Ford}
\author{Nathan F. Lepora}
\author{Efi Psomopoulou*}

\begin{affiliations}
School of Engineering Mathematics and Technology, University of Bristol, Bristol, BS8 1QU, UK \\
Email Address: efi.psomopoulou@bristol.ac.uk \\ 
\end{affiliations}


\begin{abstract}

Robotic systems increasingly demand tactile sensing that approaches the adaptability and resolution of human skin to enable dexterous manipulation and safe interaction. OptiTac is a biomimetic tactile sensor that emulates the mechanoreceptor-to-nerve architecture of human touch by pairing each mechanical pin on a soft skin with an optical fiber acting as an artificial nerve. This design demonstrates an architectural principle for routing tactile information away from the sensing surface while preserving high spatial resolution, establishing a practical route toward distributed tactile sensing in future robotic systems. By treating tactile signals as images, simple analytical methods, rather than opaque deep-learning models, are used to infer contact location, size, and shape, providing interpretable and scalable tactile intelligence. This work demonstrates how evolutionary principles from biology can guide the development of artificial nerve systems for robots, offering a pathway toward human-like tactile perception in next-generation robotic platforms. More broadly, OptiTac establishes an artificial nerve-inspired sensing framework for interpretable robotic touch and a scalable route toward future distributed tactile systems.

\end{abstract}

\section{Introduction}
Humans use the sense of touch in a wide range of applications from object identification and manipulation to interacting with the environment and other people.\textsuperscript{[1]} Tactile sensors aim to provide a ``sense of touch'' to robots to gain more complex information from the environment.\textsuperscript{[2]} With properties such as tailored tactile distribution for different areas of the skin,\textsuperscript{[3,4]} high degrees of flexibility, and hyperacuity, the human skin is a naturally-evolved tactile sensor that sets the standard for robotic systems.\textsuperscript{[5]} To improve robot dexterity, there is a need for a low-cost soft tactile sensor that provides distributed tactile sensing.\textsuperscript{[6-11]}

\begin{figure}
  \includegraphics[width=\linewidth]{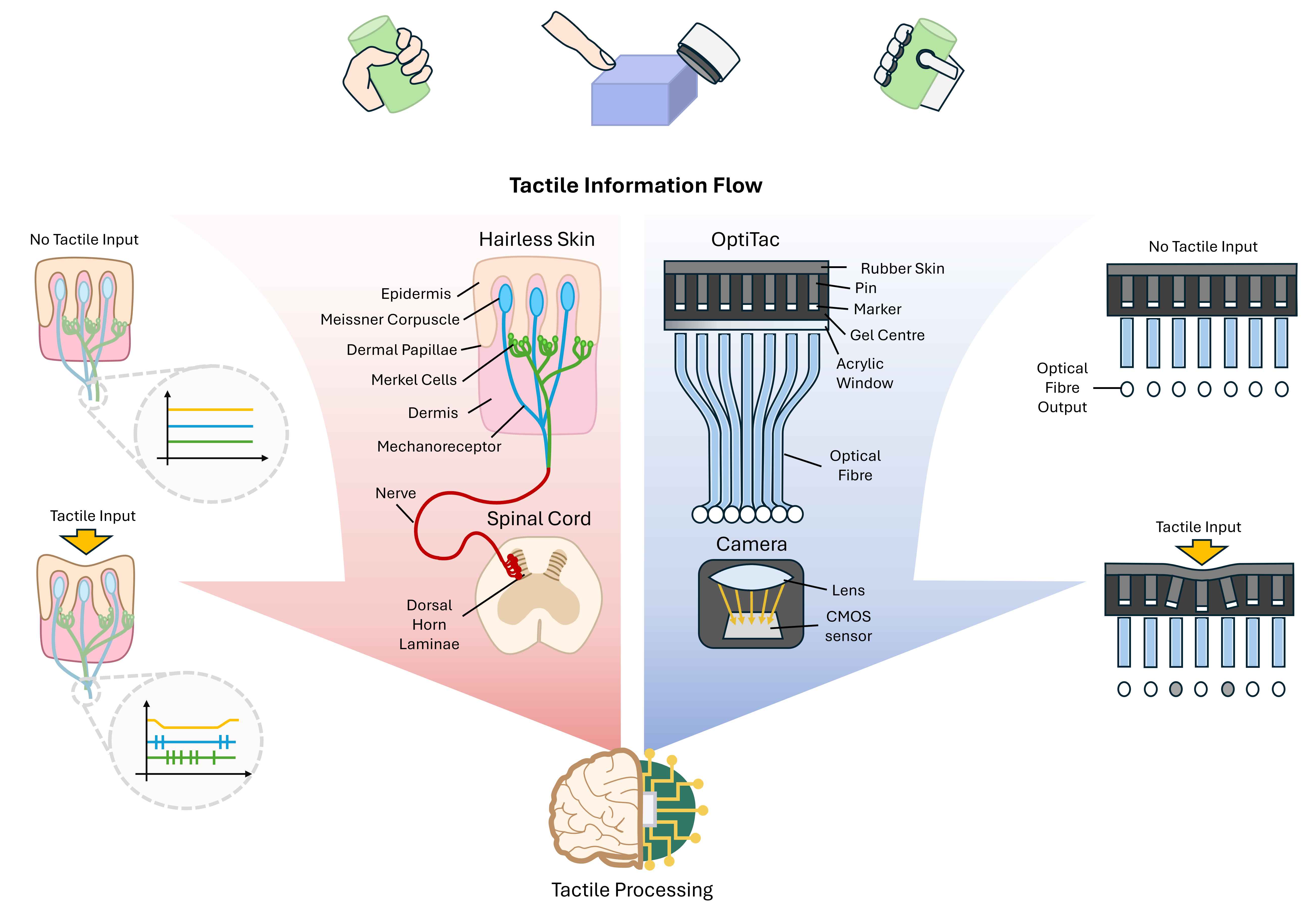}
  \vspace{-1mm}
    \caption{\textbf{Concept and biomimetic design of the OptiTac tactile sensor.} There are two subprocesses; tactile transduction and high spatial resolution sampling. Tactile transduction in the OptiTac is inspired by the way humans transduce tactile information. Hairless skin contains features such as Merkel cells and Meissner Corpuscles that are key to the transduction of mechanical stimulation of the skin and send activation signals along mechanoreceptors. Low level tactile feature extraction happens as early as the skin and the spinal cord. This is present in the OptiTac from the; TacTip skin, the optical fibers that transport and process the tactile information and the way in which the output light intensities are processed.}
    \label{fig: background}
\end{figure}

The mechanoreceptors responsible for tactile sensing use Merkel cells and Meissner Corpuscles, end organs situated within the dermis, to transduce mechanical tactile information into electrical information (\textbf{Figure~\ref{fig: background}}). Research estimates that 230,000 mechanoreceptors are distributed across the skin and that their density can range from 241,~81,~58~units/cm\(^{2}\) in the fingertips, fingers and palm, respectively, compared to 13~units/cm\(^{2}\) in arms.\textsuperscript{[12]} When manipulating objects humans use a wide range of hand grasps, from precision handling to power grips,\textsuperscript{[13]} demonstrating that not just fingertips come into contact with objects but a variety of areas across hands. Nerves transmitting tactile information away from the skin, terminate in specific areas of the spinal cord,\textsuperscript{[14,15]} this neural circuitry suggests that there is a level of tactile signal processing that occurs here, making complex representations of tactile features.\textsuperscript{[16]} Projection neurons transmit these signals to the brain, specifically the somatosensory cortex. Separate regions in the somatosensory cortex process tactile information from different regions of the skin, down to distinct regions for individual fingers.\textsuperscript{[17]} This highlights that even though there is preliminary tactile processing in the spinal cord, different areas of the skin are independently processed in the brain, which is a characteristic that could be important in developing tactile sensors. Developing a tactile sensor that processes tactile information away from the skin and can process that information from different areas of the skin could be a step toward developing distributed tactile sensing similar to that in humans.

A wide range of different tactile sensors have been developed based on piezoresistive, capacitive and optical technologies.\textsuperscript{[6]} The TacTip\textsuperscript{[18,19]} is an example of a vision-based tactile sensor (VBTS), a subset of optical tactile sensors (OTS), that uses a camera to extract high spatial resolution tactile information from the tactile skin.\textsuperscript{[20]} The tactile skin of the TacTip has a morphological pin-like structure which transduces and amplifies tactile deformations.\textsuperscript{[21]} This structure is inspired by the dermal papillae structure found in hairless human skin. Rigid pins on the skin's inner surface lever when the skin deforms. Markers on the ends of the pins highlight their position and are captured by a camera; the position and velocity of these markers mimic the signals induced by Merkel cells and Meissner Corpuscles. VBTSs have high spatial and temporal resolutions, but to increase the area of tactile skin the cameras can image, they either need large focal distances or they can require additional lenses such as fisheye lenses.\textsuperscript{[22]} Despite VBTSs achieving high levels of dexterity when integrated within robotic fingertips,\textsuperscript{[23]} the factors mentioned above limit their ability to provide distributed tactile sensing in other areas critical to achieving different tactile tasks.\textsuperscript{[24,25]} Other OTSs employ optical fibers (OFs) to transport tactile information away from the skin as they are compact, flexible and are optically isolated.\textsuperscript{[26]} Fiber bundles, a single fiber with multiple cores, have been used to produce high-resolution images of tactile skin; however, these bundles can only image small areas, so require additional lenses, and they are also restricted in their array structure due to their fabrication.\textsuperscript{[27-29]} Optical fiber arrays (OFA) consisting of individual OFs have also been successfully integrated into tactile sensors; however, these sensors use reflective/patterned tactile skins~\textsuperscript{[30-32]} or use markers that move normally to the OFA~\textsuperscript{[33]} to transduce tactile information. An early OFA tactile sensor used a design principle in which each OF represented a tactile taxel and from this a tactile sensor with a variable taxel distribution based on the distribution in the human thumb was developed.\textsuperscript{[31]} Combining the biomimetic mechanical properties of the TacTip with an OFA structure to transport tactile information away from the skin surface could be a solution toward developing human-like distributed tactile sensing for robots. 

Recent OFA tactile sensors have employed deep learning methods to extract and predict key tactile contact parameters such as; normal force, shear and torque.\textsuperscript{[32,34]} A drawback of using these methods is that they make predictions using ``black-box'' models. Treating tactile information like images is not a new method~\textsuperscript{[35]} and there are a wide range of vision-based methods that can be used to extract features from images.\textsuperscript{[36]} A simple vision-based method is to use image moments to identify local contact features such as position, orientation, and area from tactile data.\textsuperscript{[37-40]} Based on image moments, Hu moments are a set of 7 moment invariants that are used for pattern recognition.\textsuperscript{[41]} Both image moments and Hu moments have been applied to a wide range of tactile sensors, from single tactile sensors,\textsuperscript{[39,42]} to grippers with tactile sensing,\textsuperscript{[37,43]} to anthropomorphic hands with tactile finger tips and phalanges~\textsuperscript{[38]} as well as complete tactile coverage of a hand.\textsuperscript{[40]} These methods have also been used to complete high level tasks such as grasp stability of household objects.\textsuperscript{[44]} Developing an analytic protocol for extracting key contact parameters from tactile information could provide a step towards minimizing the extent at which deep learning methods are used for tactile inference. Interpretability is particularly valuable in tactile sensing because it enables contact features to be linked directly to physically meaningful deformation patterns, which can support sensor design, calibration, debugging, and transfer across sensing geometries.

We propose a generalizable design principle for distributed tactile sensing, implemented here in a high-resolution proof-of-concept prototype. The prototype is an OF based tactile sensor (termed OptiTac) which mimics the way humans transduce and transport information away from the skin, producing an artificial tactile neural pathway (Figure~\ref{fig: background}). Our tactile sensor is underpinned by a simple design principle in which each pin 
attached to the tactile skin is paired with an OF. Each OF then transports local tactile information to a camera that captures global tactile information from multiple OFs. This enables intuitive tactile information processing using image moments. Consequently, OptiTac was able to demonstrate hyperacuity, measure contact size and correctly classify various contact shapes. Note that here, we use ‘hyperacuity’ to denote localization precision substantially finer than the spatial spacing between adjacent sensing elements under the tested indentation conditions. We show that this architecture supports intuitive tactile signal formation and enables analytical inference of contact location, size, and shape, thereby providing a scalable route toward distributed tactile sensing in robots.

This paper introduces a biomimetic optical-fiber tactile architecture that creates an artificial nerve-like pathway for robotic touch; shows that one-to-one pin–fiber pairing physically structures the tactile signal for interpretable inference; and demonstrates sub-element localization, contact-size estimation, and shape classification using analytical image descriptors rather than end-to-end deep learning.

\section{Results}
In this study, a prototype OptiTac tactile sensor was built to demonstrate the performance of the biomimetic design principle. To do this, focus was placed onto developing a sensor with high spatial resolution, where OFs within the OFA were packed tightly, minimizing measurement errors of key contact parameters. By designing a high spatial resolution prototype OptiTac, we aim to test the limits of the current design principles. A key premise of this work is that biomimetic physical pre-processing at the sensor level can simplify downstream tactile inference.

\subsection{Remote signal transmission enabled by a biomimetic tactile sensor}
The OptiTac sensor design (shown in Figure~\ref{fig: background}) evolves from the standard TacTip sensor design~\textsuperscript{[21]} by separating it into two distinct modules~(\textbf{Figure~\ref{fig: sensor_structure}a}). The first is a soft tactile sensing module and the second is a remote camera module, both separated by an OFA. The sensing module is filled with transparent elastomeric gel for skin-like compliance, held in place by a transparent acrylic window (see ``Methods''). Light from an LED ring within the module evenly illuminates the white markers on the tips of pins that protrude into the gel against the black rubber skin. The light reflected by the white markers is captured by 217 plastic OFs (diameter 1~mm) held flush against the acrylic window by a 3D printed mount. The OFA is arranged in a hexagonal array~(Figure~\ref{fig: sensor_structure}b) to ensure tight packing of the OFs and coherently so that the position of one OF is known from one end of the array to the other. The light leaving the ends of the OFs is captured by a remote camera in a 3D printed casing to reduce the influence of ambient light.

Tactile deformations of the skin induce pin levering, which is highlighted by the white markers~(Figure~\ref{fig: background}). The markers move, aligning and unaligning with the OFs within the OFA. If the marker and OF are aligned, there is a high intensity light output. If there is no or little alignment, then there is a low intensity output. The output light intensities are captured by the camera which is connected to a computer, where the images are processed to reproduce high resolution light intensity surface plots (Figure~\ref{fig: sensor_structure}c). These results verify that tactile information can be transmitted from the compliant sensing interface to a physically separate imaging module through the coherent optical-fiber array while retaining a structured signal suitable for downstream contact inference under the controlled conditions examined here.

\subsection{Biomimetic one-to-one pin-OF pairing enhances intuitive feature extraction}
In the skin, mechanoreceptors can receive tactile information directly from Merkel cells or Meissner \newline corpuscles.\textsuperscript{[45]} A biomimetic one-to-one pin to OF pattern on the tactile skin was chosen to act as a physical pre-processing step to provide intuitive tactile information. We propose that simple analytical methods can be applied to characterize the contact parameters rather than using deep learning methods. 

\begin{figure}
  \includegraphics[width=\linewidth]{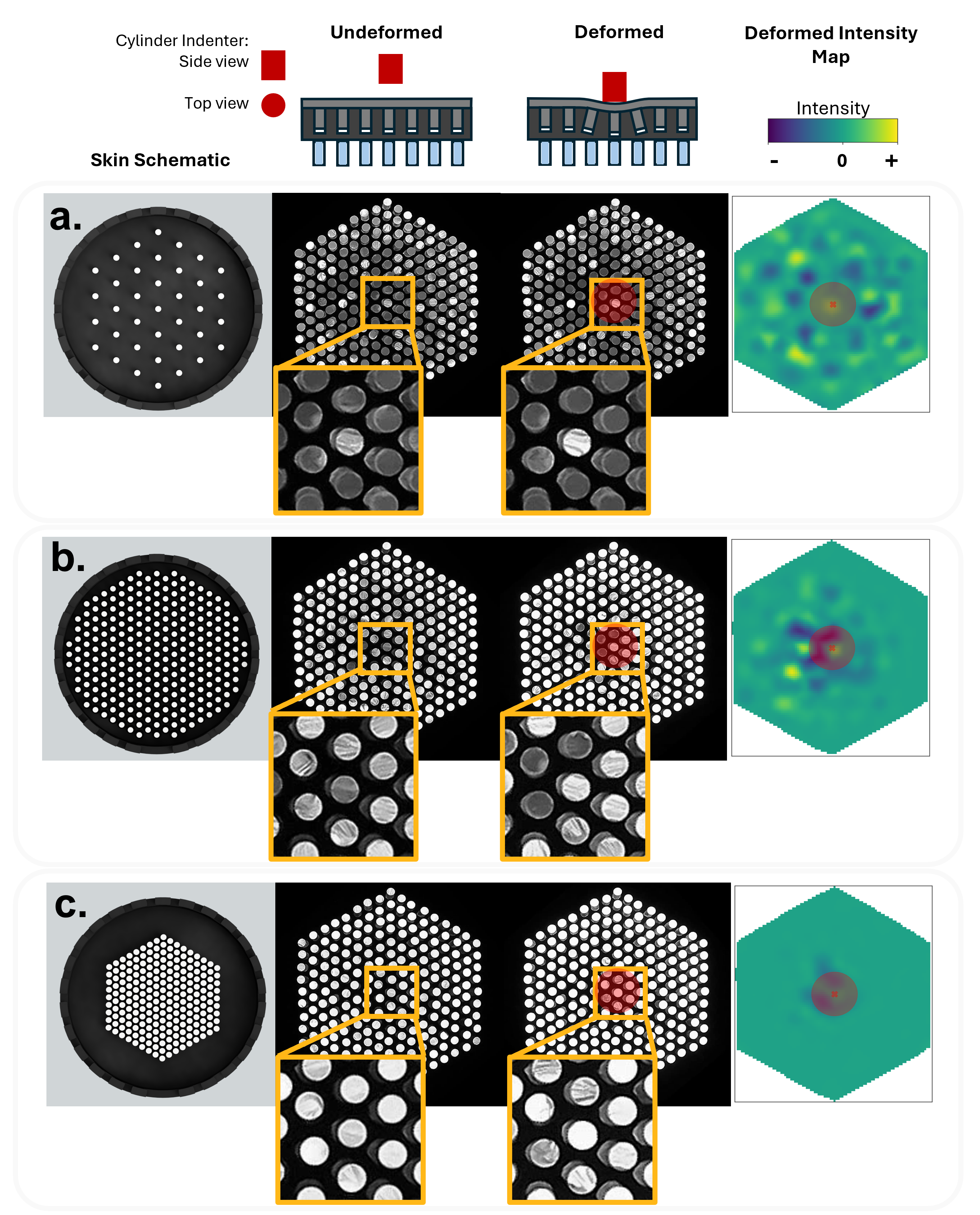}
  \caption{\textbf{Biomimetic one-to-one pin-OF pattern.} TacTip skin marker patterns \textbf{a.} sparse, \textbf{b.} dense and \textbf{c. }aligned. Left to right; CAD schematic, a cropped undeformed tactile image, a cropped tactile image under 3 mm of deformation by 5 mm diameter ball indenter and its resultant surface plot.}
    \label{fig: marker patterns}
\end{figure}

To demonstrate this, three tactile sensing modules with (i) sparse, (ii) dense, and (iii) aligned pin patterns were fabricated (see ``Methods'')~(\textbf{Figure~\ref{fig: marker patterns}}). For all designs, the diameter and length of the pins remained constant. Only the distance between the pins was changed. The camera images all OFs together, collating the local tactile information that they transmit into a global representation of the tactile skin. The sparse pin pattern (i) consists of fewer pins than the number of OFs~(Figure~\ref{fig: marker patterns}a). When there is no external object interacting with the tactile skin, the skin is undeformed and the captured tactile image shows how the sparse pattern is transmitted via the OFs to the camera. Varying levels of alignment between the pins and the OFs result in single to multiple neighboring OFs transmitting information about a single pin. The variation of skin color from low intensity in the center of the OFA to a higher intensity in the periphery is an artefact caused by the presence of the LED ring. The dense pin pattern (ii) consists of each pin being 2~mm from neighbouring pins~(Figure~\ref{fig: marker patterns}b). This distance is larger than the 1.5~mm distance between each OF and neighboring OFs. The undeformed image shows variation in the transmitted light intensity as the alignment of the pins with the OFs varies. Finally, the aligned pin pattern (iii) is the core principle of the OptiTac sensor~(Figure~\ref{fig: marker patterns}c). Here, each pin is vertically aligned with an OF and the undeformed image shows less variation in the global output light intensities compared to (ii) the dense pattern. 

The three different pin patterns on the tactile skin provide three different light intensity changes as a result of a 5~mm diameter ball indented to a depth of 3~mm. The sparse pin pattern (i) generates surface wide intensity changes~(Figure~\ref{fig: marker patterns}a). As white markers move into the field of view (FOV) of OFs that initially did not have any pin alignment, the intensity increases for those OFs. For the OFs that were initially aligned with pins that then moved out of the FOV, the intensity decreases. The dense marker pattern (ii) produced a more localized intensity change than the sparse pattern in response to the same indentation~(Figure~\ref{fig: marker patterns}b). However, this area of intensity change is still much larger than the size of the indenter, and the pattern of intensity change produced is complex and not intuitive. The aligned one-to-one pin-OF pattern (iii) that mimics the way Merkel cells and Meissner corpuscles translate tactile information to a mechanoreceptor produced a localized negative intensity change as a result of the indentation~(Figure~\ref{fig: marker patterns}c). This is an intuitive response, and analytical techniques can be applied to characterize key tactile features.

To design a tactile skin that best transmits deformations through the OF array, three different patterns were designed. Of the three, only the aligned marker pattern produced a localized intensity change. This biomimetic pairing between the pins of the tactile skin and the OFs was chosen for the OptiTac sensor.

\subsection{Simple tactile centroid localization delivers performance comparable to complex state-of-the-art methods} 
\begin{figure}
  \includegraphics[width=\linewidth]{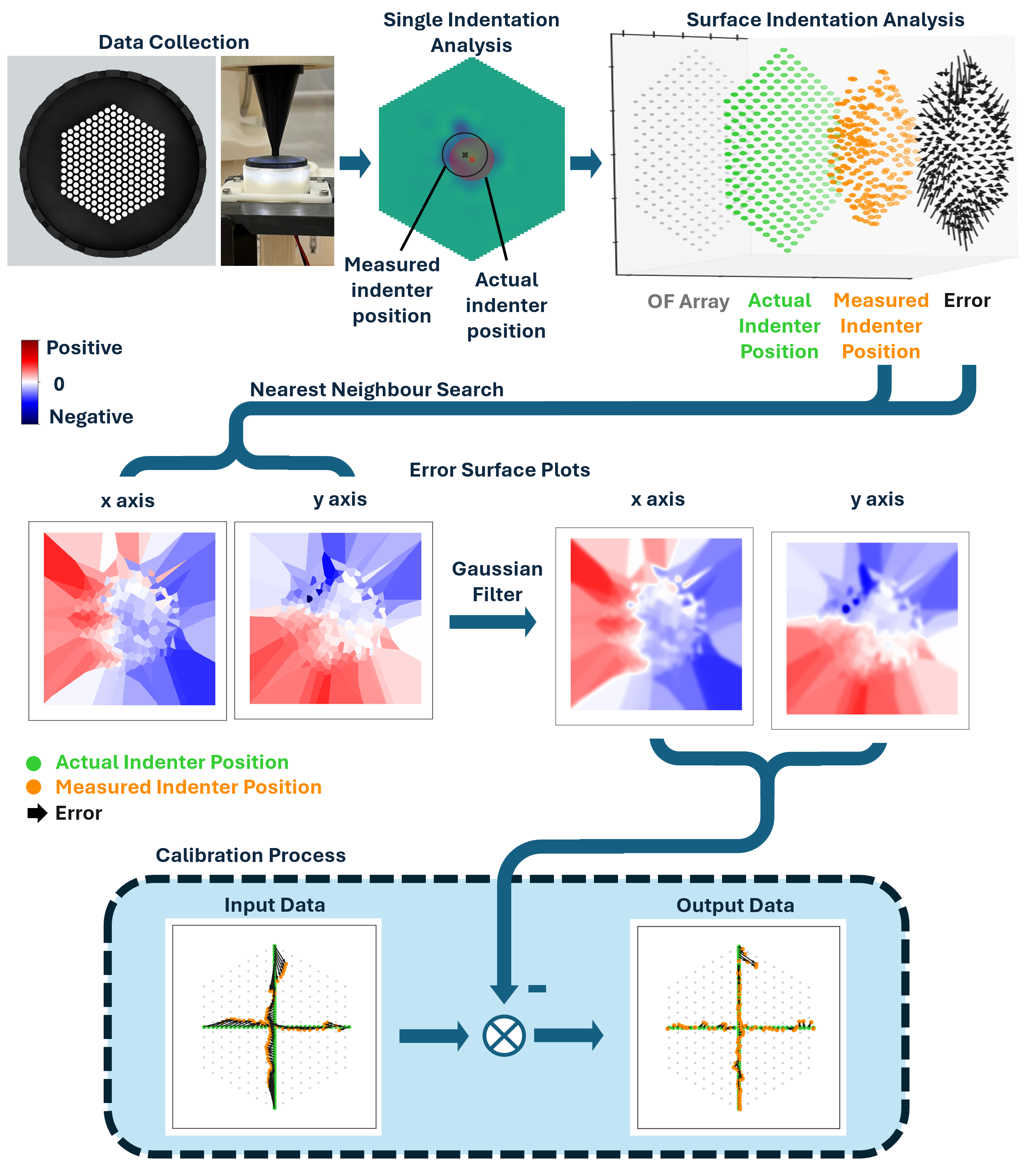}
  \caption{\textbf{Development of the contact centroid localization calibration method.} To calibrate the measured centroid values, pre-requisite information on how the errors vary over the surface of the sensor is required. Indentations are made directly above each OF within the array. The errors of the 217 indentations vary in both x and y axis so two calibration plots are generated. For each plot, a nearest neighbor search is performed on a 1000x1000 mesh assigning the error value of the nearest measured indenter position. To minimize the effect of overfitting, a Gaussian filter ($\sigma$ 10) was applied. To demonstrate, unseen input data was calibrated using these calibration surface plots.}
    \label{fig: position error calibration}
\end{figure}

Using the physical pre-processing of the tactile contact information as a result of the biomimetic one-to-one pin-OF pairing, a simple image moment method could be applied to extract contact location (see ``Methods''). To establish the accuracy of this method, indentations were made at various locations across the OptiTac. Errors between the measured centroid positions and the true indenter centroid positions were observed, which varied across the surface of the OptiTac. A calibration process was developed to reduce these errors (\textbf{Figure~\ref{fig: position error calibration}}), producing two calibration surface plots for the measured centroid position in the \(x\)- and \(y\)-axes (see ``Methods''). To evaluate the performance of the calibration method, the effect of the calibration on two data sets was assessed. The first data set was a ``seen'' data set that the calibration surface plots had been generated from. This consisted of indentations aligned with each individual OF within the OF array. The second data set was an ``unseen'' data set in which indentations were made along the \(x\)- and \(y\)-axes of the OF array. These indentations were made incrementally between the OFs within the OF array and aligned with the OFs. All indentations for both datasets were indented to a depth of 3~mm.

\begin{figure}
  \includegraphics[width=0.8\linewidth]{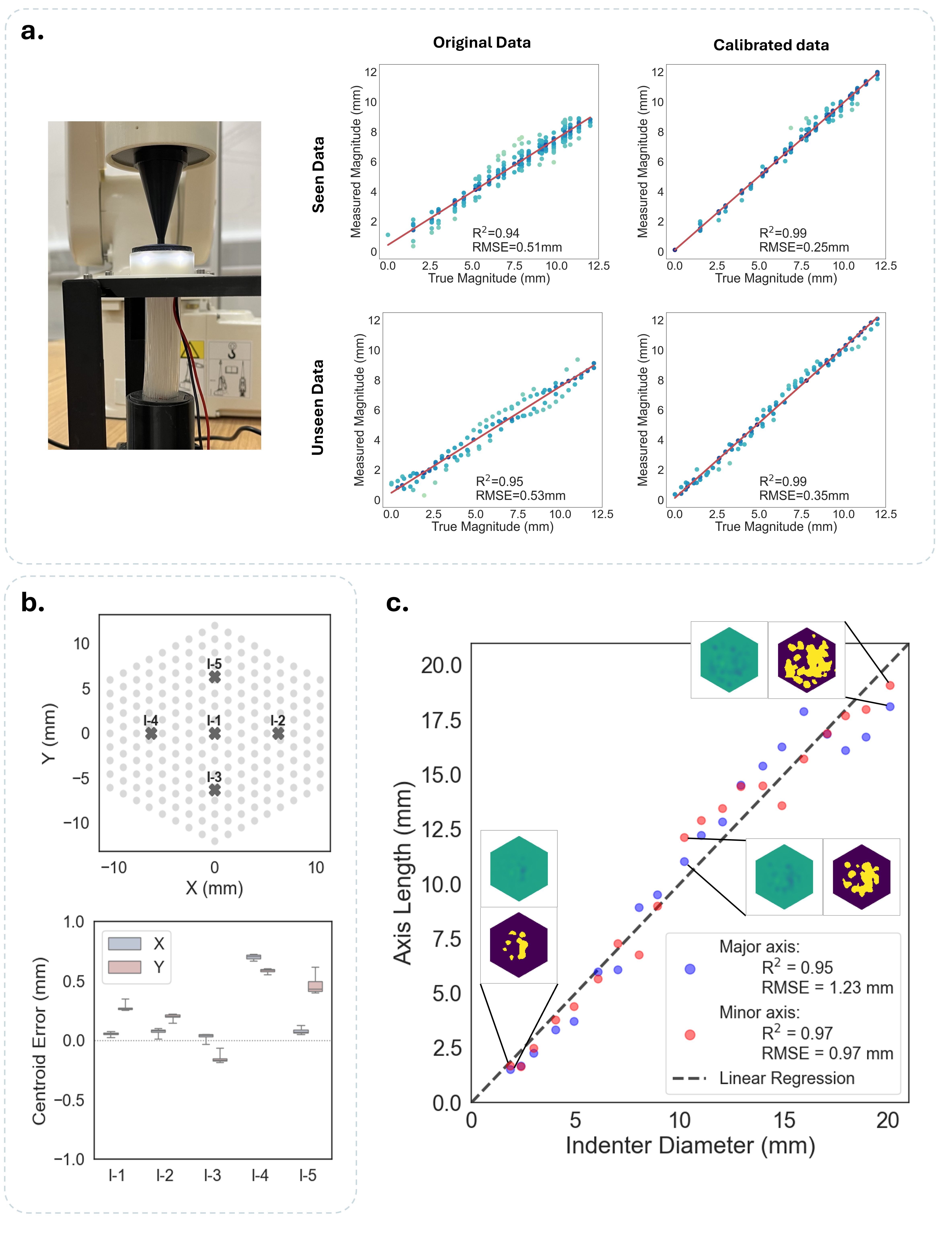}
  \caption{\textbf{Tactile contact characterization of the OptiTac.} \textbf{a.}~To demonstrate the effect of the calibration process (see ``Methods''), it was tested on seen data which was used in the development of the calibration protocol. An unseen data set which was not used in the development of the protocol was tested to demonstrate the generalizability of the calibration. The two scatter plots in the far right illustrate the effect of the calibration on the unseen data set compared to the ground truth. The gray scatter points in the background represent the position of the individual OFs within the OFA.   
    \textbf{b.}~Repeatability of the centroid measurements. Top shows the 5 ground truth positions that a 5~mm diameter circle indenter contacts the OptiTac. The gray scatter points in the background represent the position of the individual OFs within the OFA. Bottom shows the calibrated x,y centroid measurement errors at the 5 positions after a total of 10 indentations each.  \textbf{c.}~Calibrated contact width measurements along the major and minor axes. The three binarized surface plots on the right demonstrate how different indenter sizes affect the spread of the resultant light intensity changes.}
    \label{fig: centoid results}
\end{figure}

The calibration of the ``seen'' data set demonstrates how effective the error reduction is using the generated calibration surface plots~(\textbf{Figure~\ref{fig: centoid results}a}). Due to a Gaussian filter step in the process of producing the calibration surface plots, the ``seen'' data was prevented from overfitting and resulted in a calibrated \(R^{2}\) value of 0.99 and a reduction in the RMSE value by 0.26~mm. This was further demonstrated in the ability of the calibration surface plots to increase the \(R^{2}\) value to 0.99 of the ``unseen'' data. Since the ``unseen'' data also consisted of indentations between neighboring OFs, an RMSE value of 0.4~mm in both the \(x\)- and \(y\)-axes demonstrates the effectiveness of interpolating between neighboring OFs. Hyperacuity is also demonstrated here, as the degree to which the centroid of indentations can be measured is more than 3 times smaller than the distance between neighboring pin-OF pairs.

To compare with recent studies, Baimukashev, et al. [2020] developed a similar sensor using an OFA to transmit tactile information from a flat-patterned tactile skin.\textsuperscript{[32]} The spacing between 6 neighboring OFs was 3.9~mm, which is more than double the distance of the OFs in the array presented in this paper. A multi-output CNN was used to predict key contact parameters. This sensor had an RMSE value of 1.1~mm and 1.4~mm on the \(x\)- and \(y\)-axes, respectively. Lu, et al. [2023] recreated Baimukashev's sensor design and applied a Multiscale ResNet, which reduced the RMSE values to 0.03 and 0.02~mm in the \(x\)- and \(y\)-axes, respectively.\textsuperscript{[34]} OptiTac achieves sub-millimeter localization with a transparent analytical pipeline, demonstrating that physically structured tactile sensing can approach useful localization performance without requiring end-to-end deep learning.

To further verify the contact centroid method, the contact centroid was measured repeatedly at multiple locations on the OptiTac surface. The aim was to establish the range of error values and whether they differ at different locations. To assess the repeatability of the method, indentations were made 10 times to a depth of 3~mm, at five locations on the TacTip skin~(Figure~\ref{fig: centoid results}b). These indentations were performed by a 5~mm diameter flat circular indenter. Since the distance between the OFs is different along the \(x\)-axis compared to the \(y\)-axis because of the hexagonal OF array pattern, the evaluation of the centroid error was separately analyzed in both the \(x\)- and \(y\)-axes. Evaluating the repeatability of the contact centroid method revealed that the average centroid errors in the five locations were less than \(\pm\)~0.71~mm. This again demonstrates the hyperacuity property of the OptiTac. On average, at each location, the range of centroid errors was 0.09~mm, demonstrating the repeatability of the OptiTac's centroid measurements. It shows that even if a location has a large average centroid error, such as a 0.70~mm average error at position I-4, this is a repeatable error. Using the calibration method, the contact centroid can be repeatedly measured with minimal errors.

\subsection{Contact-width estimation using a simple model}
To evaluate the OptiTac's ability to accurately measure contact width (see ``Methods''), a series of different sized indenters were used to test this. Different sized circular flat indenters were pressed onto the tactile skin. These indenters were 3D printed and ranged from 1.85-20.09~mm in diameter (see ``Methods''). The indenters were circular, so it was assumed that the measured width along the major and minor axes should produce the same value. It was also assumed that the presence of no indenter would produce a 0~mm diameter measurement. An offset of 3.43~mm was removed from the measurements of the main axis and an offset of 2.47~mm was removed from the measurements of the minor axis. The measured lengths of the major and minor axes were scaled by 2.35 and 2.51, respectively, to produce the same linear response. Both the major and minor axes follow a strong linear relationship with \(R^{2}\) values of 0.95 and 0.97 respectively, indicating the model's suitability in measuring contact width over a large range of contact sizes~(Figure~\ref{fig: centoid results}c). The RMSE values for the width of the major and minor axes were 1.2~mm and 1.0~mm respectively, which are both smaller than the distance of 1.5~mm between pin-OF pairs. This shows that the interpolation step to generate the light intensity surface plots also improves the accuracy of the contact width measurements. Smaller than 10~mm in indenter diameter, the RMSE for the major and minor axes are 0.75 and 0.57~mm respectively. Indenters larger than 10~mm and the RMSE values increase to 1.52 and 1.20~mm respectively. This indicates that there are size-dependent behaviors and that the current structure of the OptiTac can measure smaller contacts with greater accuracy. Overall, the model has a strong linear relationship between the true and measured contact diameter and that the OptiTac has greater accuracy for small diameter tactile contacts. 

\subsection{Hu moments enables shape classification}\label{results: Contact Shape}
\begin{figure}
  \includegraphics[width=\linewidth]{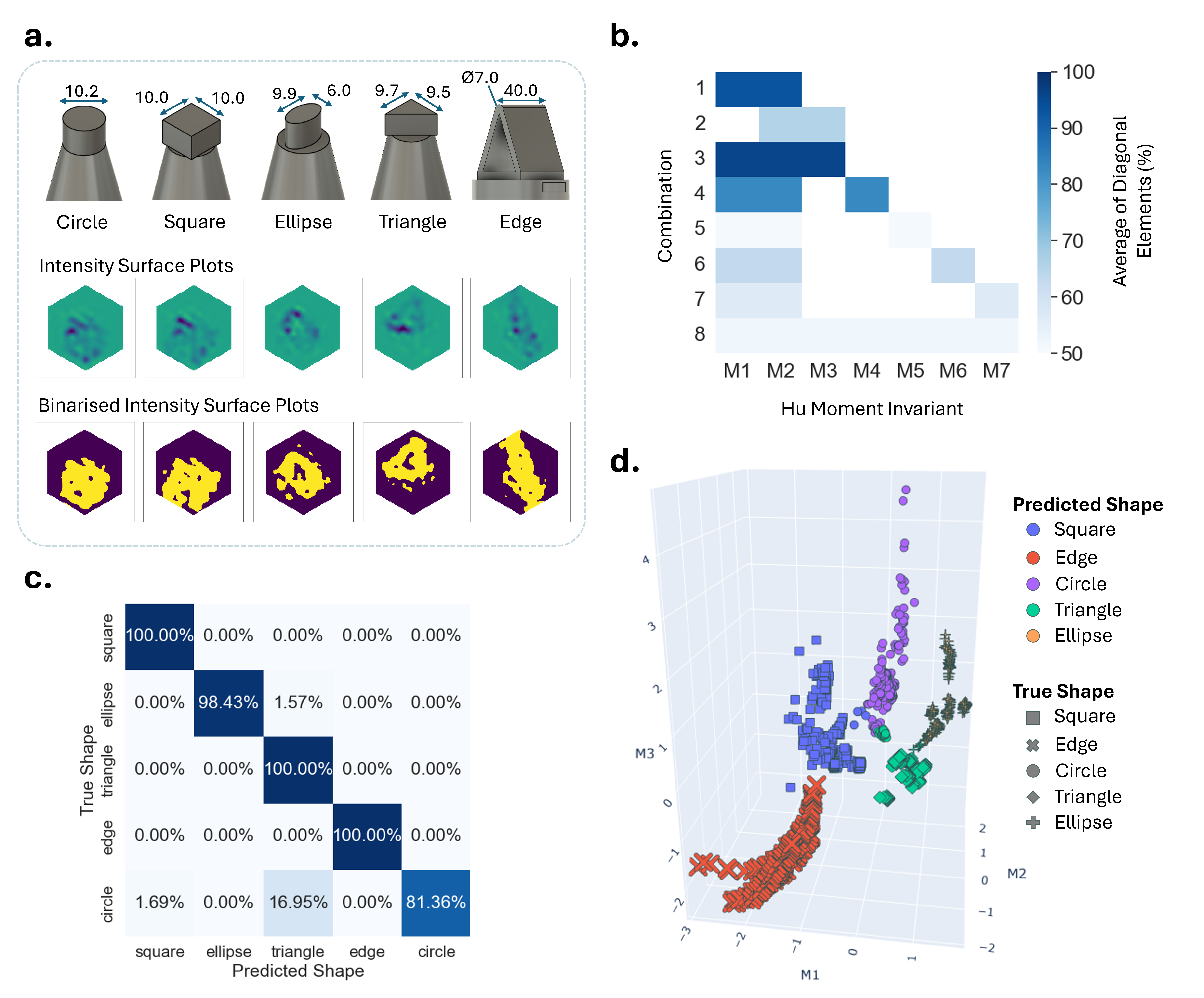}
  \caption{\textbf{Contact shape classification using the OptiTac.} \textbf{a.}~Dimensions in mm of the 3D printed indenters. An edge and four flat shapes (circle, square, ellipse, triangle). The output intensity surface plots are shown for each shape. \textbf{b.}~Hu moment combinations and the resultant average of the diagonal elements in the corresponding confusion matrices for shape classification using a GMM. \textbf{c.}~Confusion matrices for different combinations of Hu moments which are used in the Gaussian Mixture method to classify different shapes. \textbf{d.}~Clustering of data from five different shaped indenters when they are pressed into the sensor skin to a depth of 3 mm. Hu moments M1, M2 and M3 were used to extract shape features and a GMM was used to classify the clusters.}
    \label{fig: shape results}
\end{figure}

To investigate whether structured tactile signals from OptiTac support interpretable contact-shape discrimination, we evaluated the sensor on a controlled set of canonical indenters. The aim was not to solve general object recognition, but to establish proof-of-principle that simple tactile image descriptors can distinguish meaningful geometric contact classes. We therefore tested whether the sensor could first separate an edge from flat contacts and then discriminate among several representative flat shapes. A total of five 3D printed indenters were used; one edge and four flat shapes (circle, square, triangle, and ellipse)~(\textbf{Figure~\ref{fig: shape results}a}). These indenters were pressed into the OptiTac to a depth of 3~mm for 2~s, at random x,y coordinates and orientations. The coordinates were constrained so that the entire indenter remained within the bounds of the OF array. The Hu moments of the binarized intensity distributions during indentation were calculated. Different classification methods were analyzed and the Gaussian Mixture Model (GMM), an unsupervised learning method, produced the best response. A GMM uses a combination of multiple Gaussian distributions to represent the data and to estimate the probability that each data point belongs to each distribution. This was used to classify the indenter shapes according to the calculated Hu moments. The number of components of the GMM was set to 5 which is equal to the number of shaped indenters. 

To optimize the classification ability of the GMM, different combinations of Hu moments were considered. In Hu. [1962] where the set of seven Hu moments were first published, moments M1 and M2 were used to classify different alphabetical letters.\textsuperscript{[41]} It was suggested that different combinations of Hu moments could be used to produce a more optimal clustering. Applying this suggestion to the work in this paper, different combinations of Hu moments were assessed on their ability to best discriminate the different shapes. The first two combinations of moments were two dimensional like the example in Hu. [1962] (Figure~\ref{fig: shape results}b). The next five combinations were three dimensional solutions combining both M1 and M2 moments with the remaining moments respectively. The final combination was of all seven Hu moments. These eight combinations were compared to determine an optimal Hu moment combination for shape classification in the OptiTac.

To assess whether any combination of Hu moments affected the classification ability of the GMM, confusion matrices were generated.
The average of the diagonal elements of these matrices provided a metric of comparison. Only two combinations of Hu moments exceeded an average of 90~\%~(Figure~\ref{fig: shape results}b). Combining M3 with M1 and M2, increased the average value by 2.70~\% to an average of 96~\%. This suggests that important tactile features for the classification of the shapes in Figure~\ref{fig: shape results}a are present in the first three Hu moments. The subsequent Hu moments may contain tactile features, but not ones that can be used to distinguish the different shapes in this example, since there was at least a 10~\% difference in the average diagonal elements between combination 1 and combinations 4-8~(Figure~\ref{fig: shape results}b). This variation demonstrates the importance of selecting the optimal combination of Hu moments for feature extraction in shape classification tasks.

The confusion matrix for the chosen combination of Hu moments (M1, M2 and M3) breaks down the classification ability of the GMM for each shape~(Figure~\ref{fig: shape results}c). There was a 100~\% true positive classification of the edge shape. This was the only shape with 0~\% false-positive and false-negative classifications resulting in a F-1 score of 1.00. This indicates that the edge has highly distinctive features that the GMM can cluster. Among the flat shapes, both square and triangle shapes achieved a 100~\% true positive classification but had non-zero false-positive values. That means that the GMM clusters were not fully separable for the different shapes. The GMM is a statistical based clustering model, and so if shapes produce similar feature distributions, the classifications of clusters can overlap. As a result, it can be said that there are similarities in the flat shapes and the feature distributions they produce. However, all shapes had a minimum F1-score of 0.90. The chosen combination of Hu moments resulted in distinctive features clusters for each shape and enabled correct classification of an edge and different flat shapes.

To verify the classification of different extracted features, the M1, M2 and M3 values for all indentations were plotted~(Figure~\ref{fig: shape results}d). Clear clusters of data points for each of the different shapes can be observed. The GMM successfully classifies the majority of the true shape data points with minimal miss classification. Overall, by combining Hu moments M1, M2 and M3 with a GMM, the OptiTac was able to both discriminate edges from flat shapes and between the different shapes with minimal errors. These results should be interpreted as a proof-of-principle for tactile feature discrimination from physically structured contact patterns, rather than as a comprehensive benchmark for unconstrained shape recognition across arbitrary objects and interaction conditions.

\section{Discussion}
In this paper, we proposed a generalizable design principle for distributed tactile sensing and implemented it in a high-resolution proof-of-concept prototype. For this purpose we developed a novel OTS that combines the biomimetic mechanical properties of the TacTip sensor with a biomimetic OFA. We also developed an analytical methodology to process tactile images and extract useful parameters for tactile contact characterization.

The OptiTac design preserves the biomimetic mechanism of the TacTip whilst using a biomimetic OF array to transmit tactile information to a remote camera. Other tactile sensors that use OF arrays have flat reflective/patterned tactile  skin~\textsuperscript{[30-32]} or markers that move normal to the OF array~\textsuperscript{[33]} to highlight changes to the tactile skin as a result of object contact. The key feature of the TacTip is the biomimetic pins on the inner surface of the tactile skin which lever in response to tactile interactions.\textsuperscript{[21]} Choosing a one-to-one pairing between the pins and the OF array allowed intuitive feature extraction after the tactile images were pre-processed. Compared to the other pin patterns, the aligned one to one pin pattern produced only a localized negative intensity change in response to the same tactile stimulus~(Figure~\ref{fig: marker patterns}c). 
This design principle resulted in localized intensity changes as a result of deformation. This in turn allowed for intuitive feature extraction from the surface plots using image-based methods to extract key contact parameters. 
 
The OptiTac was able to extract useful contact information such as locating and determining the size of different tactile contacts and classifying the different tactile contact shapes. The calibration process successfully reduced the RMSE of the tactile contact centroid measurements with an accuracy within the sub-millimeter range. This is comparable to existing OFA tactile sensors despite using a simpler analytical approach.\textsuperscript{[32,34]} 
The RMSE values for both axes was approximately 1~mm. The errors in contact localization and width estimation might be caused by two steps in the method. The pixels that make up the endfaces of the OFs are all averaged together to produce a single value. Any spatial information that is mixed by the OFs, is lost due to the averaging. In the OFA sensors where deep learning methods are applied, the raw tactile images of the OFAs are used directly to train the models, retaining any key spatial information despite light mixing.\textsuperscript{[32,34]} The number of indentations used to make the calibration surface plots might also be a limiting factor, as the indentations are taken directly above each OF; future work will extend this to arbitrary contact distributions and broader operating conditions. Despite these limitations, the RMSE values are smaller than the spacing between the OFs that make up the array. This demonstrates the OptiTac retains the hyperacuity property of the TacTip.\textsuperscript{[21]}

The OptiTac was able to correctly classify different shapes when using a set of simple descriptors to produce a GMM rather than using deep learning models. Comparing the classification ability of different Hu moment combinations allowed for the selection of a feature vector with the highest average diagonal elements which was the first three Hu moments 96~\%. This method provides transparency in how the simple features are selected and used to develop the GMM. Deep learning methods are black boxes in terms of how and what features are selected during training of the models. The resultant GMM is interpretable as clear clusters of data points are visible~(Figure~\ref{fig: shape results}d). In this context, interpretability is not only a methodological preference but a practical advantage, because it links inferred contact properties to physically meaningful signal changes and therefore supports calibration, failure analysis, and future sensor redesign.

A key feature of OptiTac is the physical separation of the tactile sensing interface from the imaging hardware through the optical-fiber array. In the present proof-of-concept system, this separation was demonstrated under controlled laboratory conditions using relatively short fibers and fixed packaging geometry. Although the results confirm that useful tactile information can be transmitted reliably to a remote camera module, further engineering studies will be required to quantify robustness under longer fiber routing, repeated bending, packaging tolerances, and broader environmental variation. Establishing performance under these conditions will be an important next step toward deployment in distributed robotic sensing systems.

The main aim of this work was to demonstrate the performance of a proof of concept tactile sensor that preserves the biomimetic mechanical transduction properties of the TacTip with an OF array. However, the design considerations involved in the development of this prototype sensor allows the sensor to be scaled to a form factor that enables dexterous manipulation. In future work, integration with robotic end effectors, such as grippers, will be considered. To demonstrate the performance, simple feature descriptors were used to generate a GMM. To increase generalizability of the OptiTac to a wide range of objects, more advanced feature descriptors could be considered.\textsuperscript{[36]} The novel feature of the OptiTac is the OFA connecting the two modules. To assess this design in providing useful tactile information, optimal lighting was required. As a result, the LED ring from the original TacTip design was kept in the soft tactile sensing module. Future design variations could remove the LED ring to further minimize the OptiTac size.

\section{Conclusion}
In this paper, the OptiTac was presented which evolves from the standard TacTip design by integrating an OFA. This design utilizes design properties that are inspired by biological features to develop a sensor that demonstrates hyperacuity and that can accurately measure the size of tactile contacts. The image like properties of the surface plots generated by pre-processing the tactile images, enabled the classification of different contact shapes using simple image classification techniques.

\section{Methods and materials}
\subsection{Fabrication of the OptiTac tactile sensor}
To fabricate the sensing module, the flat tactile skin, pins, markers, and body were printed simultaneously using a multi-material 3D printer (Stratasys J826). The skin and pins were printed in a black rubber material (Agilus30), and the markers and body were printed in a rigid white material (VeroWhite). The flat skin was printed with a matte finish to minimize unwanted glare. A clear acrylic window, 1 mm thick, was glued to the sensor opening. A transparent elastomeric gel (TechsiL RTA27905 A/B) was injected and set, filling the cavity between the printed skin and the acrylic window. This gel adds compliance to the sensor and allows visible light to pass through. 

The OFs were cut to 100~mm in length using a blade. The aim of this paper was to prove that the OptiTac could detect key tactile information. As a result, only a short length of OF was required. Each OF endface was then manually polished sequentially with 5~\textmu m, 1~\textmu m and 0.3~\textmu m grit lapping sheets (Thorlabs). This resulted in each OF being 95~mm in length. OFs were polished normal to the lapping sheets to ensure a flat endface. A loupe was used to visually assess whether the OF faces had a smooth finish after polishing.

\subsection{Tactile image pre-processing}
\begin{figure}
  \includegraphics[width=\linewidth]{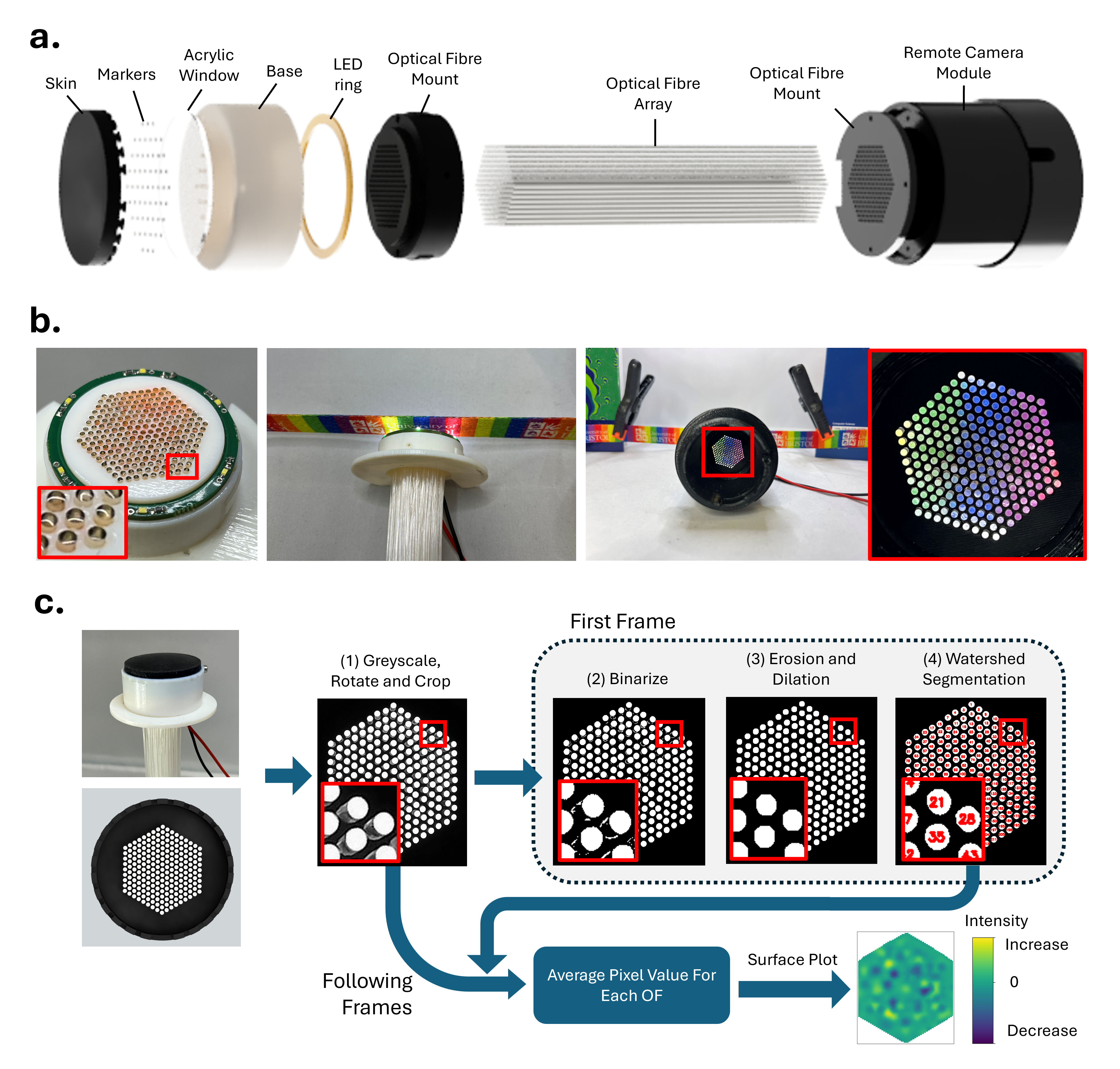}
  \caption{\textbf{Design and operating principle of the OptiTac.} \textbf{a.}~OptiTac sensor CAD schematic. \textbf{b.}~the 217 OFs are arranged in a coherent hexagonal array. To demonstrate its coherence a repeating patterned object was imaged through the array when the tactile sensing module was not attached. Clear stripes are visible at the output of the OF array. \textbf{c.}~the pre-processing of the raw tactile images to intensity surface plots from which relevant parameters can be extracted.}
  \label{fig: sensor_structure}
\end{figure}

The goal of the software methodology was to analytically measure contact parameters, such as position, width, and shape, using tactile images directly from the camera. These raw tactile images contain large amounts of redundant data as a result of the images being taken in RGB format. To remove this redundant information, pre-processing of each camera frame is required. The first stage of pre-processing each frame reduced the size of data by removing redundant information~(Figure~\ref{fig: sensor_structure}c.1). The images were first converted from a three channel RGB image to a single channel grayscale image. This was as a result of the markers on the end of the pins are white and the rubber skin being black. As the markers move, the light transmitted by the OFs changes only in intensity and not color. Following this, the frames were rotated and cropped to remove pixels that do not make up the OF array image.

The positions of the OFs were determined only in the first frame because the OFs are fixed in position and only the tactile skin changes as a result of tactile contact. This was a three-step process, where the image was binarized~(Figure~\ref{fig: sensor_structure}c.2), then the background noise was removed by erosion and dilation~(Figure~\ref{fig: sensor_structure}c.3), and finally the image was segmented using watershed segmentation~(Figure~\ref{fig: sensor_structure}c.4). Watershed segmentation provides labels for each object identified within the frame, their centroid position, and all pixels that fall within each object label. The background was given a label of 0 and ascending labels were given to each object within the image.
To minimize the effect of optical distortions induced by the lens of the camera, the pixel positions of the segmented OFs were remapped to mm positions from a CAD diagram of the 3D printed parts designed to hold the OFs in place. In this version of OptiTac, the re-mapping step had minimal effect, but this may be useful in a case where the OF array is coherent but the shape of the array at one end is different to the other. Since the OFs are fixed in position, the results of the segmentation of the first frame was applied to the subsequent frames. 

An average pixel intensity value for each OF was calculated per frame. The measured average pixel intensity was between 0-255. To highlight intensity changes from when the OptiTac was not receiving any tactile contact, an average intensity value was calculated over five frames for each OF. These values were used treated as background noise and removed from the average pixels intensity values of following frames. This process highlights any changes in the state of the OptiTac from undeformed to deformed. As markers move as a result of tactile contact, the intensity of light transmitted by the OFs changes, resulting in a positive or negative intensity change from the undeformed state.

The positions of the OFs in the tactile images can be directly linked to the positions of the OFs that capture light from the TacTip because of OF array coherence.  The OF array structure results in 217 taxels of information about the tactile skin. To improve the resolution of this, these points were interpolated to produce high resolution surface plots. From these surface plots key tactile information can be obtained. To achieve this, cubic interpolation with a mesh resolution of 100x100, was performed on the intensity values of the OF array. For each frame, a surface plot was generated to measure the changes in the intensity values of the OF array as a response to the changing marker positions. These surface plots have a greater spatial resolution than the original OF array values and are used to calculate a number of contact parameters.

\subsection{Contact Inference}
Key contact parameters can be analytically obtained by treating the surface plot like an image, using image processing techniques such as image moments. The aligned one pin to one OF pin pattern resulted in only negative intensity changes and as a result image moments can be used to quantify parameters of the contact such as position and size~(Figure~\ref{fig: marker patterns}c). 

To calculate the image moments of the surface plot, pre-processing is required. First, the intensity values are inverted so that indentations generate a positive distribution. This is because currently, tactile contacts produce a negative intensity change due to the white makers moving out of view of the OFs. Next, the inverted surface plot is truncated around a threshold to remove background noise. The surface plot was truncated rather than binarized to preserve the intensity distribution. Using image moments the following contact inferences can be made on the pre-processed surface plot.

\subsubsection{Contact Centroid} \label{method: contact centroid}
Raw image moments describe properties of a distribution relative to the origin~(\ref{eq: raw moments}). The first raw image moment can be used to describe a distribution's mean which is, in this example, the tactile contact centroid~\((\bar{x},\bar{y})\)~(\ref{eq: centroid}):
\begin{equation}\label{eq: raw moments}
    m_{i j} = \sum_{x}\sum_{y}x^{i}y^{j}I(x,y),    
\end{equation}
\begin{align}\label{eq: centroid}
    \bar{x} &= \frac{m_{1 0}}{m_{0 0}}, & \bar{y} = \frac{m_{0 1}}{m_{0 0}}.  
\end{align}

\subsubsection{Contact Width} \label{method: contact width}
The central moments are the moments of a distribution with respect to the distribution mean~(\ref{eq: central moments}). These are location-invariant moments,
\begin{equation}\label{eq: central moments}
    \mu_{i j} = \sum_{x}\sum_{y}(x-\bar{x})^{i}(y-\bar{y})^{j}I(x,y).
\end{equation}
Central moments can be used~(\ref{eq: eigen - central moments}) to calculate the indentation's eigenvalues~(\ref{eq: eigen}),
\begin{equation}\label{eq: eigen - central moments}
    \mu_{i j}^{'} = \frac{\mu_{i j}}{\mu_{0 0}},
\end{equation}
\begin{equation}\label{eq: eigen}
    \lambda_{1,2} = \frac{(\mu_{2 0}^{'}+\mu_{0 2}^{'})}{2}\pm\frac{\sqrt{(4\mu_{1 1}^{'2}) + (\mu_{2 0}^{'}-\mu_{0 2}^{'})^2}}{2}.
\end{equation}
Assuming the contact is Gaussian, these eigenvalues can then be used to calculate the width of the indentation at a specific height~(\(h\)) from the centroid~(\ref{eq:FW}). To find the full indentation width at half the mean (FWHM), \(h=2\),
\begin{equation}\label{eq:FW}
    \text{FW} = \text{principle~axes}_{1,2} =  2\sqrt{2\ln{h}}\sqrt{\lambda_{1,2}}.
\end{equation}

\subsubsection{Contact Shape Classification} \label{method: contact shape}
Using the second and third order central moments, seven orthogonal invariants can be applied and used for pattern recognition~(\ref{eq: hu moments}),\textsuperscript{[41]} 

\begin{align}
    &M1 = \mu_{20} + \mu_{02}, \notag  \\
    &M2 = (\mu_{20} - \mu_{02})^{2} + 4\mu_{11}^{2}, \notag \\
    &M3 = (\mu_{30} - 3\mu_{12})^{2} + (3\mu_{21} - \mu_{03})^{2}, \notag \\
    &M4 = (\mu_{30} + \mu_{12})^{2} + (\mu_{21} + \mu_{03})^{2}, \notag \\
    &M5 = (\mu_{30} - 3\mu_{12})(\mu_{30} + \mu_{12})\notag \left[(\mu_{30} + \mu_{12})^{2} - 3(\mu_{21} + \mu_{03})^{2} \right] \notag \\
        &\quad \quad \quad \quad  \quad  \quad  \quad  + (3\mu_{21} - \mu_{03})(\mu_{21} + \mu_{03}) \notag [3(\mu_{30} + \mu_{12})^{2} - (\mu_{21} + \mu_{03})^{2}], \notag \\
    &M6 = (\mu_{20} - \mu_{02}) \left[(\mu_{30} + \mu_{12})^{2} - (\mu_{21} + \mu_{03})^{2} \right] \notag + 4\mu_{11}(\mu_{30} + \mu_{12})(\mu_{21} + \mu_{03}),\notag  \\
    &M7 = (3\mu_{21} - \mu_{03})(\mu_{30} + \mu_{12})\notag \left[(\mu_{30} + \mu_{12})^{2} - 3(\mu_{21} + \mu_{03})^{2} \right] \notag \\
           &\quad \quad \quad \quad  \quad  \quad  \quad  - (\mu_{30} - 3\mu_{12})(\mu_{21} + \mu_{03}) \left[3(\mu_{30} + \mu_{12})^{2} - (\mu_{21} + \mu_{03})^{2} \right]. \label{eq: hu moments}
\end{align}

These were applied to a binarized version of the surface plots. A log-transform was performed on the resulting values. These moments were used identify different contact shapes. A Gaussian Mixture Model, an unsupervised learning method, was used to classify different clusters of data points.

\subsection{Calibration} \label{method: calibration}
During the initial development stages of the OptiTac, errors were observed between the measured centroid position and the true centroid position. The measured centroid positions errors were clustered towards the center of the OF array, resulting in larger errors toward the periphery of the OF array. To minimize these errors, a calibration process was proposed. 

A pre-requisite of the calibration process is knowledge of the centroid errors across the surface of the OptiTac. To obtain this, contact data was collected across the surface of the TacTip skin. Each contact was positioned directly over an individual OF within the array and all contacts were performed using the same size indenter (5~mm diameter cylinder). Together, this provided information on the variation of errors across the surface of the OptiTac.

The data of the centroid errors of the measured indenter positions are then processed to generate high resolution surface plots detailing how these errors vary across the array. First, a k-dimensional tree (k-d tree) is produced from the measured indenter positions. Second, two 1000x1000 grid meshes are created
individually for the x and y axis errors. To ensure that all possible errors are taken into account, these meshes span an area larger than the area of the OF array. A nearest-neighbor search is performed on these meshes using the k-d tree. Each mesh position is assigned an error value according to the error value at the nearest measured indenter position. Two surface plots are produced with discrete areas representing the error of the nearest measured indenter position. Finally, a Gaussian filter with a standard deviation of 10 was used to smooth the boundaries between each neighboring region and reduce the effect of overfitting. This resulted in two high resolution surface plots which describe the measured indenter position errors in both the x and y axes respectively.

When calibrating the unseen measured contact centroids, the errors found in the respective error surface plots are removed. The centroid coordinates are rounded to the nearest mesh grid value. The corresponding error values from the x axis error mesh and the y axis error mesh are subtracted from the current measured indenter position. An illustration of this calibration process on an unseen input dataset is shown in~\textbf{Figure~\ref{fig: position error calibration}}.

\subsection{Experimental Setup}
To provide accurate ground truths to measure contact position and depth, an experimental setup was established. The OptiTac was secured to a table top breadboard and all indentations were performed by an IRB 120-3/0.6 ABB robot arm. The workspace of the ABB arm was calibrated such that it was centered at the center of the OF array. The orientation of the workspace also matched the orientation of the OF array. To change the parameters of the contacts, different indenters were 3D printed and attached as end effectors to the ABB arm. The camera within the sensor had a USB connection and was connected to a PC.

\subsubsection{Contact width indenters}
Circular indenters sized from 1.85-20.09~mm were 3D printed using a Stratasys F370 printer. This size range was chosen because the quality of indenters smaller than 1.85~mm in diameter was reduced as a result of the resolution of the 3D printer. Indenters larger than 20.09~mm were not used to ensure that the local deformation caused by the indenter remained within the limits of the OF array. The indenters were indented 3.0~mm into the TacTip skin and the major and minor axes of the contact were measured.

\medskip
\textbf{Acknowledgements} \par 
This work was supported by the Horizon Europe research and innovation program under grant agreement No. 101120823 (MANiBOT).

\medskip

\textbf{References}\\

[1] A. Zimmerman, L. Bai, and D. D. Ginty, “The gentle touch receptors of mammalian skin,” Science, vol. 346, no. 6212, pp. 950–954, 2014.\\

[2] R. S. Dahiya, G. Metta, M. Valle, and G. Sandini, “Tactile sensing—from humans to humanoids,”IEEE Transactions on Robotics, vol. 26, no. 1, pp. 1–20, 2009.\\

[3] R. S. Johansson and A. B. Vallbo, “Tactile sensibility in the human hand: relative and absolute densities of four types of mechanoreceptive units in glabrous skin.,” The Journal of Physiology, vol. 286, no. 1, pp. 283–300, 1979.\\

[4] R. S. Johansson and J. R. Flanagan, “Coding and use of tactile signals from the fingertips in object manipulation tasks,” Nature Reviews Neuroscience, vol. 10, no. 5, pp. 345–359, 2009.\\

[5] J. Dargahi and S. Najarian, “Human tactile perception as a standard for artificial tactile sensing—a review,” The International Journal of Medical Robotics and Computer Assisted Surgery, vol. 1, no. 1, pp. 23–35, 2004.\\

[6] L. Zou, C. Ge, Z. J. Wang, E. Cretu, and X. Li, “Novel tactile sensor technology and smart tactile sensing systems: A review,” Sensors, vol. 17, no. 11, p. 2653, 2017.\\

[7] C. Chi, X. Sun, N. Xue, T. Li, and C. Liu, “Recent progress in technologies for tactile sensors,” Sensors, vol. 18, no. 4, p. 948, 2018.\\

[8] Z. Xia, Z. Deng, B. Fang, Y. Yang, and F. Sun, “A review on sensory perception for dexterous robotic manipulation,” International Journal of Advanced Robotic Systems, vol. 19, no. 2, 2022.\\

[9] Y. Li, P. Wang, R. Li, M. Tao, Z. Liu, and H. Qiao, “A survey of multifingered robotic manipulation: Biological results, structural evolvements, and learning methods,” Frontiers in Neurorobotics, vol. 16, p. 843267, 2022.\\

[10] S. K. Sampath, N. Wang, H. Wu, and C. Yang, “Review on human-like robot manipulation using dexterous hands.,” Cogn. Comput. Syst., vol. 5, no. 1, pp. 14–29, 2023.\\

[11] C. Wang, C. Liu, F. Shang, S. Niu, L. Ke, N. Zhang, B. Ma, R. Li, X. Sun, and S. Zhang, “Tactile sensing technology in bionic skin: A review,” Biosensors and Bioelectronics, vol. 220, p. 114882, 2023.\\

[12] G. Corniani and H. P. Saal, “Tactile innervation densities across the whole body,” Journal of Neurophysiology, vol. 124, no. 4, pp. 1229–1240, 2020.\\

[13] T. Feix, J. Romero, H.-B. Schmiedmayer, A. M. Dollar, and D. Kragic, “The GRASP taxonomy of human grasp types,” IEEE Transactions on Human-Machine Systems, vol. 46, no. 1, pp. 66–77, 2015.\\

[14] V. Abraira and D. Ginty, “The sensory neurons of touch,” Neuron, vol. 79, no. 4, pp. 618–639, 2013.\\

[15] J. Turecek, B. P. Lehnert, and D. D. Ginty, “The encoding of touch by somatotopically aligned dorsal column subdivisions,” Nature, vol. 612, no. 7939, pp. 310–315, 2022.\\

[16] A. M. Chirila, G. Rankin, S.-Y. Tseng, A. J. Emanuel, C. L. Chavez-Martinez, D. Zhang, C. D. Harvey, and D. D. Ginty, “Mechanoreceptor signal convergence and transformation in the dorsal horn flexibly shape a diversity of outputs to the brain,” Cell, vol. 185, no. 24, pp. 4541–4559, 2022.\\

[17] A. R. Sobinov and S. J. Bensmaia, “The neural mechanisms of manual dexterity,” Nature Reviews Neuroscience, vol. 22, no. 12, pp. 741–757, 2021.\\

[18] C. Chorley, C. Melhuish, T. Pipe, and J. Rossiter, “Development of a tactile sensor based on biologically inspired edge encoding,” in 2009 International Conference on Advanced Robotics, pp. 1–6, IEEE, 2009.\\

[19] B. Ward-Cherrier, N. Pestell, L. Cramphorn, B. Winstone, M. E. Giannaccini, J. Rossiter, and N. F. Lepora, “The TacTip family: Soft optical tactile sensors with 3d-printed biomimetic morphologies,” Soft Robotics, vol. 5, no. 2, pp. 216–227, 2018.\\

[20] S. Zhang, Z. Chen, Y. Gao, W. Wan, J. Shan, H. Xue, F. Sun, Y. Yang, and B. Fang, “Hardware technology of vision-based tactile sensor: A review,” IEEE Sensors Journal, vol. 22, no. 22, pp. 21410–21427, 2022.\\

[21] N. F. Lepora, “Soft biomimetic optical tactile sensing with the TacTip: A review,” IEEE Sensors Journal, vol. 21, no. 19, pp. 21131–21143, 2021.\\

[22] H. Li, Y. Lin, C. Lu, M. Yang, E. Psomopoulou, and N. F. Lepora, “Classification of vision-based tactile sensors: A review,” IEEE Sensors Journal, 2025.\\

[23] M. Yang, C. Lu, A. Church, Y. Lin, C. Ford, H. Li, E. Psomopoulou, D. A. Barton, and N. F. Lepora, “Anyrotate: Gravity-invariant in-hand object rotation with sim-to-real touch,” arXiv preprint \newline arXiv:2405.07391, 2024.\\

[24] J. Cepri\'a-Bernal and A. P\'erez-Gonz\'alez, “Dataset of tactile signatures of the human right hand in twenty-one activities of daily living using a high spatial resolution pressure sensor,” Sensors, vol. 21, no. 8, p. 2594, 2021.\\

[25] Z. Chen, H. Chen, Y. Ouyang, C. Cao, W. Gao, Q. Hu, H. Jin, and S. Zhang, “A high-resolution and whole-body dataset of hand-object contact areas based on 3d scanning method,” Scientific Data, vol. 12, no. 1, pp. 1–17, 2025.\\

[26] C. Lyu, P. Li, J. Zhang, and Y. Du, “Fiber optic sensors in tactile sensing: A review,” IEEE Transactions on Instrumentation and Measurement, 2025.\\

[27] K. L. Reichenbach and C. Xu, “Numerical analysis of light propagation in image fibers or coherent fiber bundles,” Optics Express, vol. 15, pp. 2151–2165, Mar. 2007. Publisher: Optica Publishing Group.\\

[28] Q. Li, O. Kroemer, Z. Su, F. F. Veiga, M. Kaboli, and H. J. Ritter, “A review of tactile information: Perception and action through touch,” IEEE Transactions on Robotics, vol. 36, no. 6, pp. 1619–1634, 2020.\\

[29] J. Di, Z. Dugonjic, W. Fu, T. Wu, R. Mercado, K. Sawyer, V. R. Most, G. Kammerer, S. Speidel, R. E. Fan, et al., “Using fiber optic bundles to miniaturize vision-based tactile sensors,” IEEE Transactions on Robotics, 2024.\\

[30] J. L. Schneiter and T. B. Sheridan, “An optical tactile sensor for manipulators,” Robotics and Computer-Integrated Manufacturing, vol. 1, pp. 65–71, Jan. 1984.\\

[31] S. Begej, “Planar and finger-shaped optical tactile sensors for robotic applications,” IEEE Journal on Robotics and Automation, vol. 4, pp. 472–484, Oct. 1988.\\

[32] D. Baimukashev, Z. Kappassov, and H. A. Varol, “Shear, Torsion and Pressure Tactile Sensor via Plastic Optofiber Guided Imaging,” IEEE Robotics and Automation Letters, vol. 5, pp. 2618–2625, Apr. 2020.\\

[33] J. Back, P. Dasgupta, L. Seneviratne, K. Althoefer, and H. Liu, “Feasibility study- novel optical soft tactile array sensing for minimally invasive surgery,” in 2015 IEEE/RSJ International Conference on Intelligent Robots and Systems (IROS), (Hamburg, Germany), pp. 1528–1533, IEEE, Sept. 2015.\\

[34] Z. Lu, T. Yang, Z. Cao, D. Luo, Q. Zhang, Y. Liang, and Y. Dong, “Optical soft tactile sensor algorithm based on multiscale resnet,” IEEE Sensors Journal, vol. 23, no. 10, pp. 10731–10738, 2023.\\

[35] K. J. Overton and T. Williams, “Tactile sensation for robots.,” in IJCAI, pp. 791–795, 1981.\\

[36] G. Kumar and P. K. Bhatia, “A detailed review of feature extraction in image processing systems,” in 2014 Fourth international conference on advanced computing \& communication technologies, pp. 5–12, IEEE, 2014.\\

[37] A. J. Schmid, N. Gorges, D. Goger, and H. Worn, “Opening a door with a humanoid robot using multi-sensory tactile feedback,” in 2008 IEEE international conference on robotics and automation, pp. 285–291, IEEE, 2008.\\

[38] N. Gorges, S. E. Navarro, D. G\"oger, and H. W\"orn, “Haptic object recognition using passive joints and haptic key features,” in 2010 IEEE International Conference on Robotics and Automation, pp. \newline 2349–2355, IEEE, 2010.\\

[39] Q. Li, C. Sch\"urmann, R. Haschke, and H. J. Ritter, “A control framework for tactile servoing.,” in Robotics: Science and systems, 2013.\\

[40] R. Thomasson, E. Roberge, M. R. Cutkosky, and J.-P. Roberge, “Going in blind: Object motion classification using distributed tactile sensing for safe reaching in clutter,” in 2022 IEEE/RSJ International Conference on Intelligent Robots and Systems (IROS), pp. 1440–1446, IEEE, 2022.\\

[41] M.-K. Hu, “Visual pattern recognition by moment invariants,” IRE Transactions on Information Theory, vol. 8, no. 2, pp. 179–187, 1962.\\

[42] R. A. Russell, “Object recognition by a smart tactile sensor,” in Proceedings of the Australian Conference on Robotics and Automation, pp. 93–8, 2000.\\

[43] A. Drimus, G. Kootstra, A. Bilberg, and D. Kragic, “Design of a flexible tactile sensor for classification of rigid and deformable objects,” Robotics and Autonomous Systems, vol. 62, no. 1, pp. 3–15, 2014.\\

[44] Y. Bekiroglu, J. Laaksonen, J. A. Jorgensen, V. Kyrki, and D. Kragic, “Assessing grasp stability based on learning and haptic data,” IEEE Transactions on Robotics, vol. 27, no. 3, pp. 616–629, 2011.\\

[45] A. Handler and D. D. Ginty, “The mechanosensory neurons of touch and their mechanisms of activation,” Nature Reviews Neuroscience, vol. 22, no. 9, pp. 521–537, 2021.\\

\newpage

\end{document}